# Image Reconstruction Using Deep Learning

Po-Yu Liu, 3035123887, The University of Hong Kong

Supervisor: Prof. Edmund Y. Lam

2nd examiner: Dr. Yik-Chung Wu

*Abstract*—This paper proposes a deep learning architecture that attains statistically significant improvements over traditional algorithms in Poisson image denoising espically when the noise is strong. Poisson noise commonly occurs in low-light and photon-limited settings, where the noise can be most accurately modeled by the Poission distribution. Poisson noise traditionally prevails only in specific fields such as astronomical imaging. However, with the booming market of surveillance cameras, which commonly operate in low-light environments, or mobile phones, which produce noisy night scene pictures due to lower-grade sensors, the necessity for an advanced Poisson image denoising algorithm has increased. Deep learning has achieved amazing breakthroughs in other imaging problems, such image segmentation and recognition, and this paper proposes a deep learning denoising network that outperforms traditional algorithms in Poisson denoising especially when the noise is strong. The architecture incorporates a hybrid of convolutional and deconvolutional layers along with symmetric connections. The denoising network achieved statistically significant 0.38dB, 0.68dB, and 1.04dB average PSNR gains over benchmark traditional algorithms in experiments with image peak values 4, 2, and 1. The denoising network can also operate with shorter computational time while still outperforming the benchmark algorithm by tuning the reconstruction stride sizes.

## 1 INTRODUCTION

Image reconstruction, or image restoration, refers to recovering the original clean images from corrupted ones. The corruption arises in various forms, such as motion blur, low resolution, and the topic of this paper: noise. Image noise refers the variations of color and brightness in an image with respect to an ideal image of the real scene. Image noise originates from the atmospheric disturbances [1], heat in semiconductor devices [2], or simply the stochastic process of incoming photons [3]. Visually, the noise adds "dirty" grains with random intensity to the images, which in some cases severely degrades visual pleasure and image details such as edges [4]. Image noise is ubiquitous due to lack of light, or imperfect camera sensors [5], [6].

Due to cost and space efficiency, mobiles phones are usually shipped with lower-grade camera lenses and sensors. When taking pictures, especially in the nighttime, the resulting images are usually plagued with dirty pixels, which is the image noise. With the increasing prevalence of mobiles devices, the necessity for an effective noise-removing algorithm is also increased.

Image noise can be further categorized by how it is modeled mathematically. When an image is taken in a bright setting, the Gaussian distribution can most conveniently approximate the noise characteristics [7]. The resulting noise is the called Gaussian noise, a most common and most extensively researched type of noise. Another type of noise, the Poisson noise, models the brightness of a pixel as proportional to the number of independently arriving photons. Since the photons arrive independently of each other and arrive at a fixed rate, we can assume the pixel brightness as being sampled from a Poisson distribution [8]. Poisson noise is most significant in low-light and photo-limited settings, and typical cases include night scenes, medical imaging [9] and astronomical images [10]. Poisson noise was less researched because it was far less common than Gaussian noise. But due of the booming number of smart phones, more and more Poisson-noisy night pictures are produced. Therefore, this paper only focuses on removing the Poisson noise, although the algorithm proposed in this paper might be able to adapt to other types of noise with minimal modifications.

Previous research on denoising Poisson noise includes VST+BM3D [11], [12], non-local space PCA [13], and sparse coding with dictionary learning [14]. Although these algorithms achieved varying degrees of success, a pursuit for a more effective and efficient denoising algorithm has never stopped. A popular technique in machine learning, deep learning, has been showing superior performance in other imaging problems, such as image recognition [15], [16] segmentation [17], and captioning [18], and is also a top candidate in addressing the denoising problem. Traditional algorithms are often limited in representation complexity and number of parameters. For example, sparse coding with dictionary learning assumes that each pixel can be linearly reconstructed from a sparse dictionary. Such a linearity assumption, however, oversimplifies the nature of image noise, which is often applied in a non-linear fashion. Deep learning, on the other hand, learns parameters from data without human intervention. Moreover, deep learning does not assume linearity and can learn arbitrarily complex non-linear transformation [19], [20]. For example, a deep learning model can learn a complex transformation from noisy images to clean images, and such a transformation constitutes a denoising algorithm. Burger, Schuler, and Harmeling [21] demonstrated that a simple feed-forward neural network could achieve the same level of performance as BM3D in Gaussian denoising. Other researchers also demonstrated the potential of convolutional neural network in image denoising [22], [23], [24], [25], [26]. The success stories prompted me into addressing the Poisson denoising problem with deep learning.

The major contributions of this paper can be summarized as follows:

- I proposed a deep convolutional neural network for Poisson denoising. The network incorporates two designs in other works: convolutional autoencoders [25] and symmetric connections [26]. Autoencoders [27] compress the input to compact representations via the bottleneck design. The compact representations only retain principle elements of the input while discarding insignificant information such as noise. Therefore, simple autoencoders are effective in denoising [28], and one of the variants, convolutional autoencoders, are effective in denoising images. The symmetric connections are placed between corresponding encoders and decoders. The connections are advantageous in reminding the decoders of the image details forgotten by the decoders. They also let backpropagation propagates gradients to previous layers more efficiently.
- The network contains multiple branches of convolutional autoencoders with varying depths. A deeper branch smoothes color fluctuations more effectively while sacrifices slightly more image details. By incorporating branches of varying depths, the whole network can learn color smoothing more from deeper branches while learning image details more from shallower branches.
- The network has demonstrated outstanding denoising performance compared to the non-machine learning benchmark algorithm. In a test of denoising 21 standard test images of peak values 4 with Poisson noise, the propoesed network achieved a statistically significant improvement of 0.38dB PSNR gain on average over the benchmark algorithm, and even higher gains for stronger noise.

The following sections are arranged as follows. Section 2 summarizes previous works in image denoising via both traditional and deep learning algorithms. Section 3 explains the characteristics of the Poisson denoising problem. Section 4 details the design of experiments for both the benchmark algorithm and the proposed deep learning network. Section 5 reports both the qualitative and quantitative performance of the proposed deep learning architecture, along with the effect of stride sizes and noise level on the denoising performance. Section 6 concludes that the proposed network can achieve statistically significant improvement over traditional algorithms especially when the Poisson noise strong.

## 2 RELATED WORKS

A popular non-machine learning approach [29], [30], [31] for Poisson denoising depends on variance-stabilizing transformation (VST) that transforms Poisson noisy images into Gaussian images, such as Anscombe [32] and Fisz [33], [34], and then apply effective Gaussian denosing algorithms [35], [36], [37], [38], [11] to denoise images. Among Gaussian denoising algorithms, Block-matching and 3D Filtering (BM3D) [11], a pioneering non-machine learning Gaussian denoising algorithm, is frequently employed with VST for Poisson denoising; this combination is often referred to as VST+BM3D. Other algorithms tackle Poisson noise directly without relying on Gaussian denoising algorithms, which include non-local space PCA [13], and sparse coding with dictionary learning [14].

However, the trend of tackling imaging problems with deep learning has been growing ever since deep learning achieved ground-breaking achievements in other fields such as speech recognition [39] and machine translation [40]. The deep learning approach for image denoising almost always relies on convolutional neural network (CNN) due to the network's ability to capture images' spatial locality [41]. The training styles are nevertheless split into two categories. One approach is to attempt to directly translate a noisy image to an uncorrupted image. Another approach is to extract noise from a noisy image and subtract the noisy image with the noise; this is also called residual learning [42]. Zhang, Zuo, Gu, and Zhang [23] developed a deep CNN denoiser for denoising, de-blurring, and super-resolution with residual learning. Remez et al. [24] also developed a residual CNN with a different structure that outperforms VST+BM3D and other traditional algorithms in Poisson denoising. Gondara [25] proposed a convolutional autoencoder that achieved medical image denoising. Mao, Shen, and Yang [26] improved upon a simple convolutional autoencoder and incorporated symmetric connections. They claim that the symmetric connections are valuable in combating loss of back propagation gradient and image details. Their model accomplished outstanding results in several low-level imaging problems such as denoising and super-resolution.

## 3 THE POISSON DENOISING PROBLEM

Poisson noise commonly exists in low-light, or photon-limited, images. Although the visual effect of the noise is similar to that of Gaussian noise, the properties as well as the underlying stochastic processes vary in term of mathematical modeling. This section introduces some important properties of Poisson noise, beginning with an introduction to the Poisson distribution.

### 3.1 Poisson Distribution

Poisson distribution models discrete occurrences within a time interval when the events occur in a fixed rate, and when the arrival of an event is independent of each other [8]. It is a discrete probability distribution, and its probability mass function is

$$P(k \ occurrences \ in \ interval) = e^{-\lambda} \frac{\lambda^k}{k!} \quad (1)$$

where $\lambda$ is the average number of event occurrences. This distribution is widely adopted to model events happening with a fixed rate and independently, such as the number of customers arriving at checkout counters, the number of phone calls received by a customer service center, and also the focus of this paper: the number of photons hitting the image sensors in cameras.

### 3.2 Poisson Noise

A large quantity of photons hit a sensor pixel in order to form a pixel in the resulting images. In the relatively short exposure time, which usually ranges from 0.1 to 0.01 seconds, we can assume the photons arrive at the sensor in a fixed rate. We can also assume that photons arrive independently. Therefore, the number of photons hitting the sensor, and hence the brightness, can be approximated by a Poisson distribution. If $\lambda$ is very large, as is usually the case in picture taking, the Poisson distribution resembles a Gaussian distribution. The prevalence is why Gaussian noise is the most investigated in image denoising. Nevertheless, when the pictures are taken in low-light settings, the $\lambda$ will be small, and the Gaussian approximation no longer holds true. The Poisson noise thus explains a major proportion of the noise in low-light images. More specifically, let $X$ modeling the number of photons in a noisy pixel, $Y$ modeling the number of photons in a clean pixel. Then $X$ and $Y$ will be related by the following formula

$$P(X = n | Y = \lambda) = \begin{cases} e^{-\lambda} \frac{\lambda^k}{k!} & \lambda > 0 \\ \delta_n & \lambda = 0 \end{cases} \quad (2)$$

where $\delta_n$ denotes a probability distribution with 0 appearing with probability of 1. Unlike Gaussian noise, where the noise is determined by a single parameter: the standard deviation, Poisson noise is not determined by any specific parameter except for the pixel intensity as $\lambda$. As a result, the peak value of $Y$ is conventionally adopted to define the strength of Poisson noise in an image [24]. Lower peak values such as 1 and 2 result in stronger Poisson noise, while higher peak values such as 8 and 16 result in weaker Poisson noise. The rationale behind this convention will be discussed in Section 3.3.

### 3.3 Signal-to-noise Ratio

The true signal's power divided by the noise's power, namely the signal-to-noise ratio, is a commonly embraced measure to quantify the strength of noise. When the ratio is higher, the image enjoys a higher quality. For Poisson noise, assuming the average photon numbers in a pixel is $Y$, and the noise level is defined by the standard deviation, which is $\sqrt{Y}$, the signal-to-noise ratio is roughly

$$\frac{Y}{\sqrt{Y}} = \sqrt{Y} \quad (3)$$

As the brightness becomes higher, the signal-to-noise ratio also becomes higher, so that a Poisson noisy image will be cleaner compared to low-light images. This corresponds to our previous statement that Poisson noise is most significant in low-light settings. We can also observe from this equation that when the peak value of Y becomes higher, the noise becomes weaker and vice versa. Therefore, when we apply the Poisson noise to an image, the peak value of the image is used to control the strength of the applied Poisson noise..

However, the above statement is only useful for controlling the level of applied noise. To precisely quantify a denoising algorithm's effectiveness, the peak signal-to-noise ratio (PSNR) is commonly adopted. Its definition is

$$10 \ log_{10} \left( \frac{MAX_I^2}{MSE} \right) \quad (4)$$

where $MSE$ stands for mean squared error, and $MAX_I$ stands for the greatest potential pixel intensity in image $I$, 255 in the case of an 8-bit grey scale image. A higher PSNR in a reconstructed image represents a cleaner reconstruction compared with the uncorrupted image.

# 4 DESIGN OF EXPERIMENTS

This section presents details regarding the data sources, and the experiment designs for both the traditional benchmark and the proposed deep learning algorithms.

## 4.1 Data

I chose the PASCAL Visual Object Classes 2010 (PASCAL VOC) images for deep learning training, which contains a total of 11,321 images, and I also used a set of 21 standard test images for final evaluation and visual impression. Standard test images are a small set of conventional images widely used across various image processing works to produce comparable results, which include well-known images such as pepper and Lena. All the images are turned to grey scale for ease of experimentation. To generate a image training dataset suitable for deep learning training and prediction, I extracted 64 image patches of dimension 64×64 from each of the PASCAL VOC image as the training data for the deep learning network. Any larger patch size would be too memory consuming and infeasible be fed into my computer's memory all at once.

## 4.2 VST+BM3D

Block-matching and 3D Filtering (BM3D) with Variance-stabilizing transformation (VST) [12] is a very popular non-machine learning Poisson denoising algorithm, and I adopted it as the non-machine learning benchmark algorithm in this paper. BM3D is a pioneering non-machine learning Gaussian denoising algorithm [11]. To denoise a patch of a larger image, this algorithm first searches other resembling patches. These patches are transformed into a 3D matrix, filtered in the 3D space, and the transformed back to 2D space to obtain the denoised patch. This algorithm is designed for Gaussian noise specifically. Therefore, when we perform Poisson denoising through BM3D, VST is applied to a Poisson noisy image before applying BM3D so that the noisy image fed into BM3D will be approximately Gaussian noisy.

VST is a subclass of data transformations in applied statistics. The transformation is a function $f$ where values $x$ in a dataset are fed as the input to generate values $y$ where

$$y = f(x)$$

so that the $y$'s variance is fixed and independent of their mean values. A specific kind of transformation, Anscombe transform [32], converts a random variable of Poission distribution to approximately standard Gaussian distribution by making the standard deviation approximately constant at 1. The formula of this transformation is

$$y = 2\sqrt{x + \frac{3}{8}} \quad (5)$$

where $x$ are the original values and $y$ are the transformed values. After applying BM3D on the transformed values $y$, we conduct the inverse transform and obtain the denoised image. The inverse transform can be done by simply reversing the role of $x$ and $y$ so that

$$x = \left(\frac{y}{2}\right)^2 - \frac{3}{8} \quad (6)$$

However, Mäkitalo and Foi [30] indicates that this naive inverse transform will bias the estimators. To eliminate the bias, the following closed-form approximation of an unbiased transform

$$x = \frac{1}{4}y^2 - \frac{1}{8} + \frac{1}{4}\sqrt{\frac{3}{2}}y^{-1} - \frac{11}{8}y^{-2} + \frac{5}{8}\sqrt{\frac{3}{2}}y^{-3} \quad (7)$$

should be used instead, and this is the inverse transform employed in this paper.

## 4.3 Deep Learning Approach

Figure 1 visualizes the deep learning denoising network proposed in this paper for Poisson denoising. Arrows marked "Conv" are convolutional layers, arrows marked "Deconv" are deconvolutional layers, and arrows without any marks are simple connections. The 3D blocks represent the input or output tensors of the neural layers.

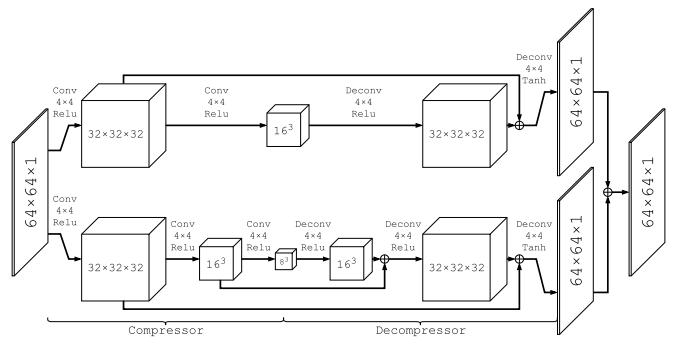

*Figure 1.* ***Visualization of the proposed deep learning architecture****. This figure illustrates the deep learning architecture proposed in this paper. This network contains 2 branches. The lower branch contains 3 convolutional layers appended by 3 deconvolutional layers, and the upper branch contains 2 convolutional layers appended by 2 deconvolutional layers.*

The network contains two branches, and each branch contains two main components: the "compressor" built with convolutional layers, followed by the "decompressor" built with deconvolutional layers. A noisy image patch of 64×64 is fed into the input, and the corresponding clean patch of the same size is fed into the output to instruct the network how to transform a noisy patch into a clean one. As an example, when a noisy patch is passed through the upper branch, the first convolutional layer transforms it into 32 smaller images of 32×32, and the second convolutional layer further compresses the 32 images into 16 smaller images of 16×16. The following 2 deconvolutional layers reverse the operations and reconstruct the smaller images back to the original size. Because the convolutional layers compress a large patch into images of lower resolutions, only representative components of the image will be retained, while less representative components such as noise will be removed. The function of the convolutional layers is also similar to a down-sampler. The compact representation is then passed through a series of deconvolutional layers, which reconstruct the representation back to a 64×64 denoised image patch. The deconvolutional layers act as an up-sampler in the process.

However, although the compressor-decompressior hybrid architecture suppresses noise, it inevitably degrades some of the image details such as object edges in the lossy compression process. In order to mitigate this problem, the network contains two branches with different compression ratios. The lower branch possesses a higher compression ratio, so that it can suppress more noise but retain less details. The upper branch possesses a lower compression ratio, so that it can retain more details but suppress less noise. By incorporating both branch into a single network, the network can learn how to suppress noise from the lower branch while learning how to recover the image details from the upper branch.

In addition, the network is also special in the symmetric connections between each pair of convolutional and deconvolutional layers with the same dimension. These connections mitigate two major problems common in training a simple network [26]. First, the connections alleviate the loss of details during the compression process. By "reminding" the later layers of the noisy but uncompressed images at the previous layers, the network can reconstruct principle features from the compressed images and reconstruct details from the noisy but uncompressed images. Second, it is generally more difficult to propagate gradients back to bottom layers in a deep network. With the connections, the gradients would be more effectively and efficiently propagated in the backpropagation process.

This network has a total of 39,098 parameters to be trained. With a training sample of 579,635 patches and validation sample of 144,909 patches, RMSProp as the optimizer, mean squared error as the cost function, and 100 as the training batch, the deep learning architecture took 2,000 seconds for a single training epoch on an Nvidia Quadro K620 GPU. I trained one network for each noise level, and each network took around 40 epochs to finish training. I chose the mean squared error (MSE) as the loss function because my evaluation metric, PSNR, links directly to the mean squared error. Minimizing MSE is equivalent to maximizing PSNR.

When denoising an image larger than 64×64, overlapping patches of 64×64 are extracted from the image with a fixed stride, and the patches are denoised individually. To reconstruct the clean image from the clean patches, the patches are stacked back to the original positions, and overlapping regions are averaged by 2D Gaussian weights. For example, when the reconstruction stride is 2, an image of 512×512 yields $\left((512-64)/2 + 1\right)^2 = 50{,}625$ patches. Each of the patches is denoised by the network before stacked back to the original position. A smaller reconstruction stride size yields more patches, which results in longer computational time but a more accurate reconstruction. While larger stride yields less patches, which results in shorter computation time but a less accurate reconstruction.

## 5 EMPIRICAL RESULTS

This section presents the empirical results, both qualitatively and quantitatively and the effect of the reconstruction stride size and noise level on the denoising performance, for the denoising network proposed in Section 4.3.

### 5.1 Visual Impression

Figure 2 visualizes the denoised images for both the deep learning algorithm with stride 1 and the benchmark algorithm VST+BM3D. The images are a sample of the standard test images, and the noisy images are clean images applied with Poisson noise with image peak value 4. The resulting PSNR values are reported on top of each noisy image; the higher PSNR, the cleaner the image is.

Numerically, the proposed denoising network performs denoising more accurately than the benchmark algorithm by achieving higher PSNRs. Visually, in the 1st set of images, we can observe that the woman's cheek and chin are plagued by uneven color fluctuations when denoised by VST+BM3D, while the same locations are smoother when denoised by the denoising network. The background also demonstrates similar effects, where my denoising network delivers smoother color transitions. The denoising network also exhibits similar advantages in the 2nd set of images, where surfaces of the peppers are smoother with less color fluctuations and ripples. The lusters on the peppers are also more accurately recovered.

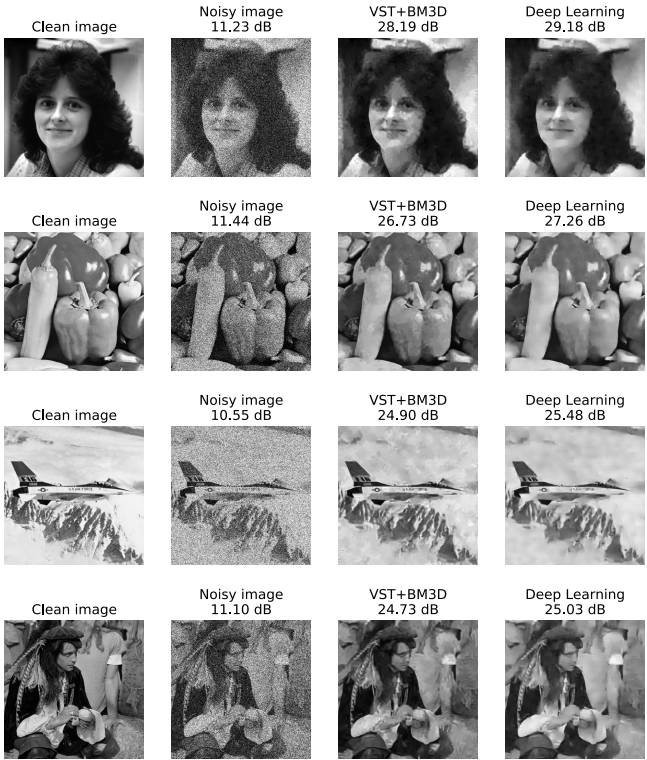

*Figure 2. **Visual impression of denoising algorithms**. This figure presents the visual impression and resulting PSNR values for both the deep learning and benchmark denoising algorithm. The noisy images are obtained by applying Poisson noise when the clean images are of peak value 4.*

Moreover, the denoising network does not simply smooth out the color fluctuations at the cost of details. In the 3[rd] set of images, the "F-16" word on the back wing of the fighter jet is as clear when denoised by the denoising network as by VST+BM3D. In the 4[th] set of images, the stripes on the hat's feather when denoised by the denoising network are also at the same clarity level compared to the benchmark. Roughly speaking, the denoising network possesses a stronger smoothing power than VST+BM3D without severely sacrificing image details, and therefore can denoise with better performance than the benchmark.

### 5.2    Quantitative Comparisons

Table 1 reports the individual PSNR values for each of the 21 standard test images denoised with the denoising network with stride 1 and the benchmark algorithm when the noise is Poisson noise with image peak value 4. The winning PSNR values between the two algorithms are marked bold.

| Image ID | 1 | 2 | 3 | 4 | 5 | 6 | 7 | 8 | 9 | 10 | 11 |
|---|---|---|---|---|---|---|---|---|---|---|---|
| (1) VST+BM3D | 24.14 | 26.78 | 25.37 | 28.34 | 24.98 | 27.76 | **21.00** | 24.89 | 21.95 | 23.84 | 26.30 |
| (2) Deep Learning | **24.63** | **27.32** | **26.34** | **28.75** | **25.31** | **28.70** | 20.86 | **25.48** | **22.25** | **24.16** | **26.64** |
| (2) - (1) | 0.49 | 0.54 | 0.96 | 0.40 | 0.33 | 0.93 | -0.14 | 0.58 | 0.30 | 0.32 | 0.34 |

| 12 | 13 | 14 | 15 | 16 | 17 | 18 | 19 | 20 | 21 | Mean | | |
|---|---|---|---|---|---|---|---|---|---|---|---|---|
| **23.82** | 23.99 | **22.28** | 26.73 | 24.73 | 26.16 | 21.98 | 25.27 | 28.13 | 28.36 | 25.09 | | |
| 23.00 | **24.30** | 22.13 | **27.26** | **25.03** | **26.84** | **22.25** | **25.48** | **29.18** | **28.86** | 25.47 | t stat | p-value |
| -0.82 | 0.32 | -0.16 | 0.53 | 0.30 | 0.68 | 0.27 | 0.21 | 1.05 | 0.50 | 0.38 | 4.2418 | 0.0004 |

*Table 1. **PSNR for each denoised standard test image**. This table presents the PSNR (dB) for each denoised standard test image with each denoising algorithm when the noise is Poisson noise of image peak value 4. Winning PSNR values are marked bold. The last row reports individual PSNR gains of the denoising network compared with the benchmark algorithm. This table also reports the t-test results against the null hypothesis that the average improvement is 0.*

The denoising network outperforms the benchmark algorithm in 18 of the 21 test images, and achieved an average PSNR gain of 0.38dB. To further test whether the improvement is statistically significant enough, a two-tail t-test over the PSNR improvements is conducted against the null hypothesis that the average PSNR gain is 0. The resulting t statistic is 4.2418, and the p-value is 0.0004. I can thus reject the null hypothesis under a significance level of 0.05 and assert that my denoising network achieves statistically significant improvements over the benchmark algorithm VST+BM3D.

### 5.3    Effect of Stride Size

While the denoising network delivers superior performance compared to VST+BM3D, its execution time is approximately 20 times longer than VST+BM3D when the reconstruction stride is 1. An immediate approach to shorten the computational time is by enlarging the stride so that the patch numbers are decreased. However, reducing the quantity of patches will inevitably degrade the denoising accuracy. Table 2 reports the effect of stride size on both the computational time per image and the average PSNR gain in the 21 test images with Poisson noise with image peak value 4.

| | VST+BM3D | Deep Learning | | | | | |
|---|---|---|---|---|---|---|---|
| Stride | | 1 | 2 | 4 | 8 | 16 | 32 |
| Time per Image (s) | 7.74 | 131.23 | 31.57 | 8.03 | 2.08 | 0.56 | 0.16 |
| Average PSNR | 25.09 | 25.47 | 25.43 | 25.36 | 25.32 | 25.32 | 25.32 |
| PSNR Gain | | 0.38 | 0.35 | 0.28 | 0.23 | 0.23 | 0.23 |
| t stat | | 4.2418 | 3.9187 | 3.2585 | 2.8097 | 2.8097 | 2.8089 |
| p-value | | 0.0004 | 0.0009 | 0.0039 | 0.0108 | 0.0108 | 0.0108 |

*Table 2. **Effect of stride size on computational time and denoising accuracy**. This table reports the effect of the stride size on the computational time and denoising accuracy. As the stride size increases, the computational time is significantly decreased, while the average PSNR gain is only slightly affected.*

Theoretically, whenever the stride size is doubled, the number of patches is reduced by 4 times, and so is the computational time. The empirical data reported in Table 2 supports this idea, where the computational time per images is 131 seconds with reconstruction stride 1, and decreased by roughly 4 times whenever the stride size is doubled. Starting from stride size of 8 up to 32, the denoising network runs faster than VST+BM3D while maintaining statistically significant PSNR gains. In practice, when we adopt the denoising network as a solution toward image denoising, a tradeoff between computational speed and denoising accuracy should be considered. In real-time applications where low latency is desired, we should choose large stride sizes for fastest possible computational time while keeping a statistically significant PSNR gain. When the reconstruction accuracy is desired, we should choose smaller stride size so that an optimal reconstruction is achieved.

## 5.4  Effect of Noise Strength

Section 5.1 to section 5.3 focus on the denoising network's performance when the images are Poisson noisy with image peak value 4. This section further analyzes the denoising network's performance under different noise levels. I trained one network for each of tested peak values 1, 2, 8 and 16 in addition to the original peak value 4. Poisson noise with peak value 1 and 2 are stronger than peak value 4, specifically when the peak value is 1, the noisy images consist of only 2 to 3 brightness levels, which is a severe degradation to the original image. Poisson noise with peak value 8 and 16, on the other hand, are weaker noise compared to peak value 4. Table 3 reports the performance of the denoising network under various peak values when the reconstruction stride is 1. It can be observed that when the noise is weak, as in the cases of peak value 8 and 16, the denoising network only wins by a small margin or even loses compared to the benchmark algorithm. But when the noise level is strong, as in the cases of peak values 1 and 2, the denoising network demonstrates superior denoising accuracy with statistically significant PSNR gains of 1.04 and 0.68 respectively. In addition, the denoising network wins 100% and 95.24% of the cases in the 21 standard test images when the peak values are 1 and 2.

| Peak Value | 1 | 2 | 4 | 8 | 16 |
|---|---|---|---|---|---|
| VST+BM3D PSNR | 21.92 | 23.56 | 25.09 | 26.69 | **28.23** |
| Deep Learning PSNR | **22.97** | **24.24** | **25.47** | **26.73** | 28.12 |
| PSNR Gain | 1.04 | 0.68 | 0.38 | 0.04 | -0.11 |
| Win | 100.00% | 95.24% | 85.71% | 71.43% | 33.33% |
| t stat | 9.52 | 7.57 | 4.24 | 0.68 | -1.71 |
| p-value | 7.17E-09 | 2.68E-07 | 4.00E-04 | 5.05E-01 | 1.02E-01 |

*Table 3. **Deep learning denoising performance under different noise levels**. This table reports the deep learning denoising performance compared with the benchmark algorithm VST+BM3D under various noise levels. The "Win" row reports the percentage of the 21 standard test images where the deep learning achieves higher PSNR. The winning PSNR values are marked bold, and the results for t test against the null hypothesis that PSNR equals zero are reported. The reconstruction strides are 1 for all cases in this table.*

Figure 3 and Figure 4 further visualizes the images reconstructed by the denoising network compared to the benchmark algorithm. As the noise becomes stronger, the images reconstructed by VST+BM3D are plagued by stronger color fluctuations in large color chunks. The denoising network, on the other hand, succeeds in obtaining cleaner color chunks without blurring object edges. Even when the peak value is 1, where the noise is so strong to the extent that the noisy images only contain 2 to 3 grayscale levels, the denoising network still successfully smoothes the large color chunks while being able to retain clear object edges.

It can thus be concluded that the denoising network demonstrates a general tendency to be more superior against the benchmark algorithm when the noise is stronger. The reasons behind this phenomenon, although not verified in this paper, could stem from two possible factors. The first reason might be that when the noise is weak, the compression power of the denoising network is too strong so that after the noise is removed, some of the image details are also removed, while the VST+BM3D could adapt to weaker noise and maintain as many image details as possible. The second reason might be that although VST can transform a Poisson distribution to a Gaussian distribution with a constant variance of 1, this relationship is most strongly held when the mean of the transformed Poisson distribution is larger than 4 [30]. When this mean value drops below 4, the variance of the Gaussian distribution begins to drop below 1 and become unstable. BM3D needs to know the noise variance in advance before denoising Gaussian noise. When the variance is unknown, as commonly in the cases for Poisson noise of peak value 1 and 2, BM3D performs suboptimally.

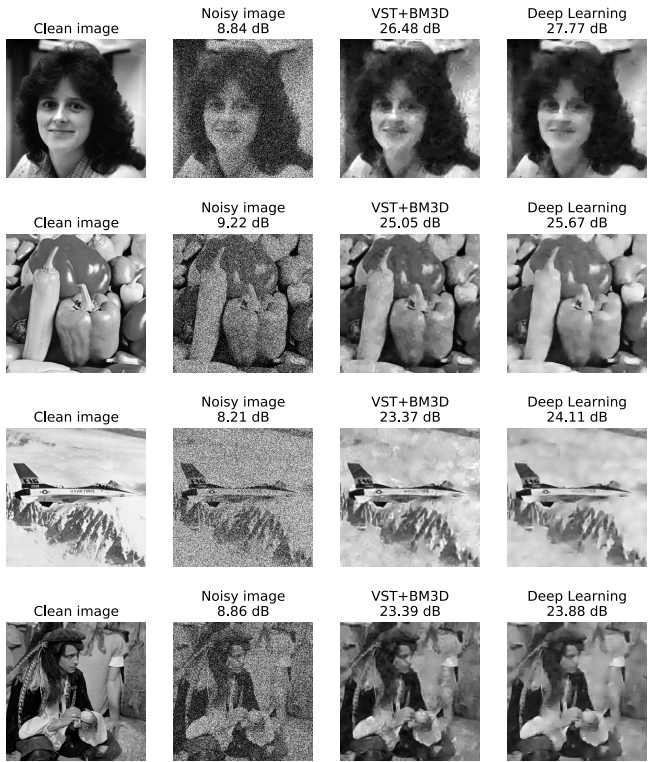

*Figure 3. Visual impression of denoising algorithms with image peak value 2.*

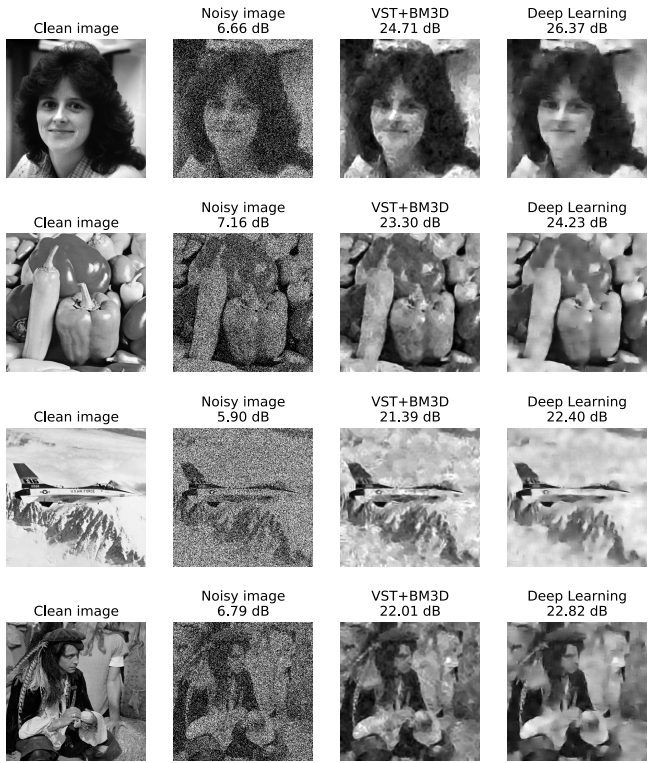

*Figure 4. Visual impression of denoising algorithms with image peak value 1.*

# 6 CONCLUSION

Although state-of-the-art non-machine learning algorithms for image denoising exist, we are constantly wondering that can we achieve better performance with the assistance of deep learning. This paper proposes a deep learning denoising network that achieves statistically significant improvements over traditional benchmark algorithms in Poisson denoising. The denoising network can achieve a statistically significant 0.38dB PSNR gain under Poisson noise of peak value 4, and even more superior PSNR gain when the noise is stronger, 1.04dB and 0.68dB for peak values 1 and 2, respectively. Although the denoising network is 20 times slower than the benchmark algorithm in computational speed with reconstruction stride 1, by tuning the stride size, the network can achieve a faster computational speed while maintaining a positive PSNR gain. Due to the computer's computational capability, this network contains only 6 layers, which is relatively shallow compared to modern architectures [42], [43] where more than 50 or 100 layers are made feasible. Nevertheless, this network performs Poisson denoising without being specifically taught the noise characteristics, while still being able to learn the parameters from the data alone. Future research could investigate whether the network can also learn to denoise other types of noise, such as the Gaussian noise or a random noise with unknown characteristics. Future research can investigate whether this architecture can be adapted to other imaging problems such as deblurring or impainting.

# 7 REFERENCES

[1] A. K. Boyat and B. K. Joshi, "A Review Paper: Noise Models in Digital Image Processing," *Signal & Image Processing : An International Journal,* vol. 6, no. 2, pp. 63-75, 2015.

[2] M. Covington, Digital SLR Astrophotography, Cambridge University Press, 2007.

[3] H. Cao, Y. Ling, J. Y. Xu, C. Cao and P. Kumar, "Photon Statistics of Random Lasers with Resonant Feedback," *Physical Review Letters,* vol. 86, no. 20, pp. 4524-4527, 2001.

[4] R. M. Willett and R. D. Nowak, "Platelets: a multiscale approach for recovering edges and surfaces in photon-limited medical imaging," *IEEE Transactions on Medical Imaging,* vol. 22, no. 3, pp. 332-350, 2003.

[5] K. Irie, A. E. McKinnon, K. Unsworth and I. Woodhead, "A model for measurement of noise in CCD digital-video cameras,"

*Measurement Science and Technology,* vol. 19, no. 4, 2008.

[6] J. Pawley, Handbook of Biological Confocal Microscopy, Springer, 2006.

[7] A. Foi, M. Trimeche, V. Katkovnik and K. Egiazarian, "Practical Poissonian-Gaussian noise modeling and fitting for single-image raw-data," *IEEE Transactions on Image Processing,* vol. 17, no. 10, pp. 1737-1754, 2008.

[8] F. A. Haight, Handbook of the Poisson Distribution, John Wiley & Sons, 1967.

[9] D. L. Snyder and M. I. Miller, Random point processes in time and space, Springer Science & Business Media, 2012.

[10] J. Schmitt, J. Fadili and I. A. Grenier, "Poisson denoising on the sphere: application to the Fermi gamma ray space telescope," *Astronomy and Astrophysics,* vol. 517, 2010.

[11] K. Dabov, A. Foi, V. Katkovnik and K. O. Egiazarian, "Image denoising with block-matching and 3D filtering," in *Image Processing: Algorithms and Systems, Neural Networks, and Machine Learning*, vol. 6064, International Society for Optics and Photonics, 2006, pp. 354-365.

[12] L. Azzari and A. Foi, "Variance Stabilization for Noisy+Estimate Combination in Iterative Poisson Denoising," *IEEE Signal Processing Letters,* vol. 23, no. 8, pp. 1086-1090, 2016.

[13] J. Salmon, Z. Harmany, C.-A. Deledalle and R. Willett, "Poisson noise reduction with non-local PCA," *Journal of mathematical imaging and vision,* vol. 48, no. 2, pp. 279-294, 2014.

[14] R. Giryes and M. Elad, "Sparsity Based Poisson Denoising with Dictionary Learning," *IEEE Transactions on Image Processing,* vol. 23, no. 12, pp. 5057-5069, 2014.

[15] Y. LeCun, B. E. Boser, J. S. Denker, D. Henderson, R. E. Howard, W. E. Hubbard and L. D. Jackel, "Handwritten digit recognition with a back-propagation network," in *Advances in neural information processing systems*, 1990, pp. 396-404.

[16] A. Krizhevsky, I. Sutskever and G. E. Hinton, "ImageNet classification with deep convolutional neural networks," in *Advances in neural information processing systems*, 2012, pp. 1097-1105.

[17] O. Ronneberger, P. Fischer and T. Brox, "U-net: Convolutional networks for biomedical image segmentation," in *International Conference on Medical image computing and computer-assisted intervention*, 2015, pp. 234-241.

[18] K. Xu, J. Ba, R. Kiros, K. Cho, A. C. Courville, R. Salakhudinov, R. Zemel and Y. Bengio, "Show, attend and tell: Neural image caption generation with visual attention," in *International Conference on Machine Learning*, 2015, pp. 2048-2057.

[19] K. Hornik, M. Stinchcombe and H. White, "Multilayer feedforward networks are universal approximators," *Neural Networks,* vol. 2, no. 5, pp. 359-366, 1989.

[20] Y. Cho and L. K. Saul, "Large-margin classification in infinite neural networks," *Neural Computation,* vol. 22, no. 10, pp. 2678-2697, 2010.

[21] H. C. Burger, C. J. Schuler and S. Harmeling, "Image denoising: Can plain neural networks compete with BM3D?," in *IEEE Conference on Computer Vision and Pattern Recognition (CVPR)*, 2012, pp. 2392-2399.

[22] L. Xu, J. S. J. Ren, C. Liu and J. Jia, "Deep convolutional neural network for image deconvolution," in *Advances in Neural Information Processing Systems*, 2014, pp. 1790-1798.

[23] K. Zhang, W. Zuo, S. Gu and L. Zhang, "Learning deep CNN denoiser prior for image restoration," *arXiv preprint,* 2017.

[24] T. Remez, O. Litany, R. Giryes and A. M. Bronstein, "Deep Convolutional Denoising of Low-Light Images," *arXiv preprint,* vol. arXiv:1701.01687, 2017.

[25] L. Gondara, "Medical image denoising using convolutional denoising autoencoders," in *IEEE 16th International Conference on Data Mining Workshops (ICDMW)*, 2016, pp. 241-246.

[26] X.-J. Mao, C. Shen and Y.-B. Yang, "Image restoration using convolutional auto-encoders with symmetric skip connections," *arXiv preprint,* vol. abs/1606.08921, 2016.

[27] D. E. Rumelhart, G. E. Hinton and R. J. Williams, "Learning representations by back-propagating errors," *Nature,* vol. 323, no. 6088, pp. 533-536, 1986.

[28] K. G. Lore, A. Akintayo and S. Sarkar, "LLNet: A deep autoencoder approach to natural low-light image enhancement," *Pattern Recognition,* vol. 61, pp. 650-662, 2017.

[29] F.-X. Dupé, J. M. Fadili and J.-L. Starck, "A Proximal Iteration for Deconvolving Poisson Noisy Images Using Sparse Representations,"

*IEEE Transactions on Image Processing,* vol. 18, no. 2, pp. 310-321, 2009.

[30] M. Mäkitalo and A. Foi, "Optimal Inversion of the Anscombe Transformation in Low-Count Poisson Image Denoising," *IEEE Transactions on Image Processing,* vol. 20, no. 1, pp. 99-109, 2011.

[31] B. Zhang, J. M. Fadili and J.-L. Starck, "Wavelets, ridgelets, and curvelets for Poisson noise removal," *IEEE Transactions on Image Processing,* vol. 17, no. 7, pp. 1093-1108, 2008.

[32] F. J. Anscombe, "The transformation of Poisson, binomial and negative-binomial data," *Biometrika,* vol. 35, no. 3/4, pp. 246-254, 1948.

[33] M. Fisz, "The Limiting Distribution of a Function of Two Independent Random Variables and its Statistical Application," *Colloquium Mathematicum,* vol. 3, no. 2, p. 138–146, 1955.

[34] P. Fryzlewicz and G. P. Nason, "A Haar-Fisz Algorithm for Poisson Intensity Estimation," *Journal of Computational and Graphical Statistics,* vol. 13, no. 3, pp. 621-638, 2004.

[35] W. Dong, G. Shi, Y. Ma and X. Li, "Image Restoration via Simultaneous Sparse Coding: Where Structured Sparsity Meets Gaussian Scale Mixture," *International Journal of Computer Vision,* vol. 114, no. 2-3, pp. 217-232, 2015.

[36] W. Dong, L. Zhang, G. Shi and X. Li, "Nonlocally Centralized Sparse Representation for Image Restoration," *IEEE Transactions on Image Processing,* vol. 22, no. 4, pp. 1620-1630, 2013.

[37] J. Mairal, F. R. Bach, J. Ponce, G. Sapiro and A. Zisserman, "Non-local sparse models for image restoration," in *2009 IEEE 12th International Conference on Computer Vision*, 2009, pp. 2272-2279.

[38] U. Schmidt, Q. Gao and S. Roth, "A generative perspective on MRFs in low-level vision," in *IEEE Conference on Computer Vision and Pattern Recognition (CVPR)*, 2010, pp. 1751-1758.

[39] A. Graves, A.-r. Mohamed and G. E. Hinton, "Speech recognition with deep recurrent neural networks," in *IEEE International Conference on Acoustics, Speech and Signal Processing (ICASSP)*, 2013, pp. 6645-6649.

[40] Y. LeCun, Y. Bengio and G. Hinton, "Deep learning," *Nature,* vol. 521, no. 7553, p. 436, 2015.

[41] K. Fukushima, "Neocognitron: A Self-organizing Neural Network Model for a Mechanism of Pattern Recognition Unaffected by Shift in Position," *Biological Cybernetics,* vol. 36, no. 4, pp. 193-202, 1980.

[42] K. He, X. Zhang, S. Ren and J. Sun, "Deep Residual Learning for Image Recognition," in *Proceedings of the IEEE conference on computer vision and pattern recognition,* 2016, pp. 770-778.

[43] K. He, X. Zhang, S. Ren and J. Sun, "Identity Mappings in Deep Residual Networks," in *European Conference on Computer Vision,* 2016, pp. 630-645.

[44] D. L. Snyder, A. M. Hammoud and R. L. White, "Image recovery from data acquired with a charge-coupled-device camera," *Journal of the Optical Society of America A,* vol. 10, no. 5, pp. 1014-1023, 1993.